\documentclass{bmvc2k}
\usepackage{gensymb}
\usepackage{amssymb}
\usepackage{wrapfig}
\usepackage{bbm}
\usepackage{makecell}
\usepackage{graphicx}

\title{Neighbourhood-Insensitive Point Cloud Normal Estimation Network}

\addauthor{Zirui Wang}{ryan@robots.ox.ac.uk}{1}
\addauthor{Victor Adrian Prisacariu}{victor@robots.ox.ac.uk}{1}

\addinstitution{
 Active Vision Lab, \\
 University of Oxford, \\
 Oxford, UK
}

\runninghead{Z. Wang and V. Prisacariu}{Neighbourhood-Insensitive Normal}

\begin{document}

\maketitle

\begin{abstract}
We introduce a novel self-attention-based normal estimation network that is able to focus softly on relevant points and adjust the softness by learning a temperature parameter, making it able to work naturally and effectively within a large neighbourhood range. As a result, our model outperforms all existing normal estimation algorithms by a large margin, achieving 94.1\% accuracy in comparison with the previous state of the art of 91.2\%, with a 25x smaller model and 12x faster inference time. We also use point-to-plane Iterative Closest Point (ICP) as an application case to show that our normal estimations lead to faster convergence than normal estimations from other methods, without manually fine-tuning neighbourhood range parameters. Code available at \url{https://code.active.vision}.

\end{abstract}

\section{Introduction}
\label{sec:intro}

Normal estimation is an important low-level task that forms the foundation for many high-level applications, such as tracking, reconstruction, and rendering. Principle Component Analysis (PCA)~\cite{hoppe1992surface}, the classic normal estimation method, estimates a normal vector at a point by collecting its $k$ nearest neighbours ($k$-NN) and fitting a plane on them. It is worth noting that, the neighbourhood range $k$, is a global hyper-parameter requiring manual tuning for PCA to achieve best accuracy. In practice, a single global $k$ is often not suitable to an entire scene, where different regions have different geometric structures. Therefore, an algorithm that can determine the neighbourhood range adaptively based on local geometric structures and is insensitive to the neighbourhood variance is desirable, as downstream tasks can enjoy high-quality normal estimations without manually tuning neighbourhood range parameters.

One possible approach to achieve this is to identify appropriate neighbourhood scales to estimate normals under different geometric properties. One line of research~\cite{guerrero2018pcpnet, ben2019nesti} aims to solve this problem by integrating information from multi-scale patches. Here, the experts that specialise at small scales are responsible for sharper regions and the experts that specialise at large scales are responsible for smoother regions. A scale manager can choose to weight estimations from different experts. This approach is effective and does not require a manual fine-tuning of the neighbourhood range, but it comes at the expenses of a large neural network, i.e. for a model that processes $L$ scales, the model is $L$ times as large as a single scale model, for example, the 7-expert Nesti-Net~\cite{ben2019nesti} model is about 1.1GB.

In this paper, we propose a normal estimation network that selects neighbour points softly according to its local geometric properties, by applying a multi-head self-attention module to the neighbour points around the current point. Our model allows a soft selection of neighbour points by learning a temperature parameter that controls the softmax function within the attention module. As a result, our method does not need a carefully picked $k$. Given a fairly large neighbourhood range, for example, setting $k$ to 30, 40, or even 50 when the best $k$ for PCA is 8, our network is able to focus on relevant points adaptively and outperforms all existing normal estimation algorithms consistently, with a significantly smaller model and faster inference time. We also provide a point-to-plane ICP \cite{chen1992object} application study to demonstrate how downstream algorithms like ICP can benefit from our neighbourhood-insensitive normal estimation.

The rest of the paper is organised as follows: Sec.~\ref{sec:related_work} introduces related works on point cloud reasoning, normal estimation, and attention mechanisms. Sec.~\ref{sec:methods} describes our network in detail. We present our results in Sec.~\ref{sec:experiments}, together with a point-to-plane ICP application study, and conclude in Sec.~\ref{sec:conclusion}.

\section{Related Work}\label{sec:related_work}
In this section, we present approaches related to our work, starting with neural network point cloud reasoning, moving to normal estimation from point clouds and ending with selected attention-based methods.

\textbf{Neural Network Point Cloud Reasoning}
Learning-based 3D point cloud reasoning is an active research area. Su et al.~\cite{su2015multi} first proposed to process point clouds using CNNs on their 2D projections. This method can extract semantic information but lose geometric details significantly. Qi et al.~\cite{qi2016volumetric} proposed to analyse point clouds using volumetric representations so that 3D CNNs could be applied directly. This method preserves 3D geometric structures but has an expensive computational cost. Geometric details are also lost during the volumetric quantization. To overcome these issues, PointNet~\cite{PointNet_Charles2017} was proposed to consume point clouds directly. The follow-up work PointNet++~\cite{PointNet2_Qi2017} improves the performance using a hierarchical strategy. More recent works, such as DGCNN~\cite{DGCNN_Wang2018}, PointCNN~\cite{PointCNN_Li2018}, and KPConv~\cite{thomas2019kpconv} combine graph neural networks and deformable convolutions with point cloud processing. These approaches focus on standard tasks like classification and segmentation. The next section focus on approaches for normal estimation.

\textbf{Point Cloud Surface Normal Estimation}
PCA~\cite{hoppe1992surface} is a classic method for normal estimation. Given a target point and its neighbours, PCA treats these points as samples from a multivariate distribution, of which the three spatial coordinates $x, y, z$ are three independent random variables. By applying eigen-decomposition to the covariance matrix of this point cloud patch, the eigenvector that has the smallest eigenvalue points to the normal direction or its reverse direction. Jet~\cite{cazals2005estimating_jet} estimates a normal vector by fitting a truncated Taylor expansion to the local point cloud patch. Geometric properties such as surface normal and the two principle curvatures can be computed from the Taylor expression.

Recently, learning-based point cloud normal estimation approaches have been proposed. PCPNet~\cite{guerrero2018pcpnet} applies PointNet to a local patch and regresses a surface normal directly. To adapt to local geometries better, it also introduces a multi-scale model that runs PointNet~\cite{PointNet_Charles2017} at three different local neighbourhood scales. Similarly, Nesti-Net~\cite{ben2019nesti} is a Mix of Expert (MoE) \cite{hampshire1992meta,jacobs1991adaptive} system that processes 7 different scales of neighbours around a target point. Each expert learns to specialise at regions with certain geometric properties and outputs confidences at inference time. To obtain the final result, a manager network integrates normal estimations from all experts and outputs a final result. Both methods tackle normal estimation by exploring more scale levels and balancing results in the multi-scale inferences. This line of approaches achieves better performance than classical PCA and Jet but also requires a large model and a long inference time. The multi-scale setting also prohibits information flow among sub-networks. Our method improves both performance and efficiency by using a single network that learns to select point information softly and can adapt to different geometries without quantizing a local patch to a number of scales.

\textbf{Attention Mechanism}
Attention was first introduced by Bahdanau et al.~\cite{bahdanau2015neural} to handle long-range dependencies in language processing tasks together with Recurrent Neural Networks (RNN). Self-attention or intra-attention~\cite{cheng2016long, lin2017structured} was proposed to understand relationships within a sequence itself. Vaswani et al.~\cite{vaswani2017attention} designed multi-head attention and replaced RNNs with self-attention to enable parallel processing, which is difficult to achieve in RNNs. Recently, attention mechanisms have been introduced to the point cloud domain to explore relationships within a point cloud. These methods use attention mechanism to enhance 
semantic understanding over the entire point cloud. \cite{yang2019modeling, yan2020pointASNL} uses attention mechanism to sample points. Wang et al. \cite{wang2019graph} employs self-attention in semantic segmentation tasks. DCP \cite{DCP} applies attention mechanism to find correspondences between two point clouds in a registration task. Inspired by prior work, we propose applying attention to normal estimation, a task that requires an understanding of local geometric structures.

\section{Method}\label{sec:methods}
Given a point cloud, our model predicts a surface normal for every point within a point cloud. To achieve this, our model takes $k$ nearest neighbours around a point, which we call a local patch, as input and outputs the predicted surface normal for this local patch as a three dimensional unit vector. The rest of the section introduces the overall normal estimation pipeline, elaborates our key attention module, and provides implementation details in the end.

    \begin{figure}[b]
        \centering
        \includegraphics[width=0.95\linewidth]{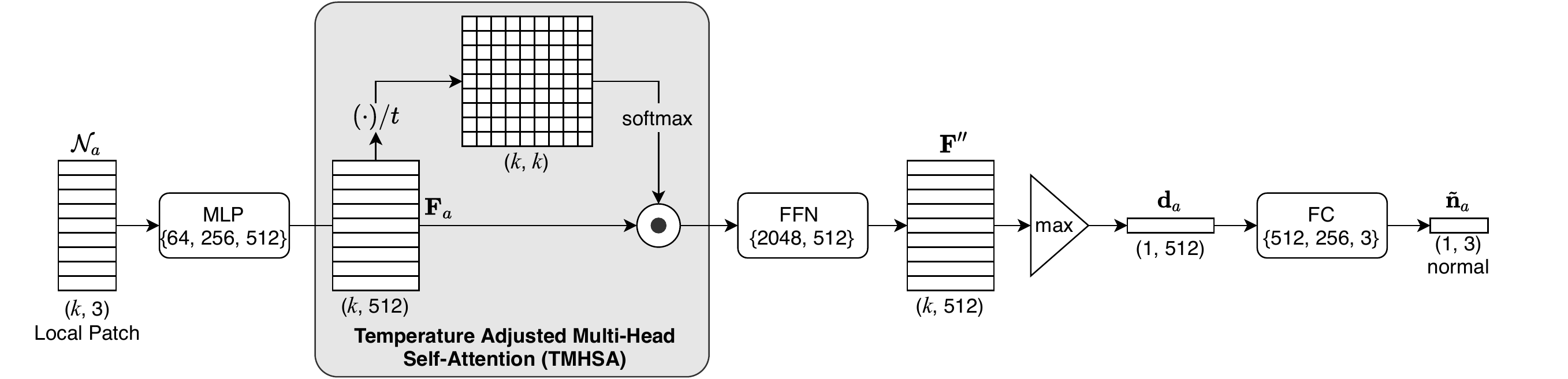}
        \caption{Our attention-based normal estimation model. We introduce TMHSA module that enables our network to focus on relevant neighbours, adapt to various geometric structures, and adjust the attention strength, given a large neighbourhood range. The MLP denotes the Multi-Layer Perceptron and the FFN denotes the Feed Forward Network introduced in \cite{vaswani2017attention}.}
        \label{fig:method_pipeline}
    \end{figure}

\textbf{Pipeline}
Our prediction pipeline consists of five steps:
    \textbf{1)} The network takes a local patch $\mathcal{N}_a$ at a point $a$ as input and extract a set of per-point features $\mathbf{F}_a := \{f_i, i \in \mathcal{N}_a\}, \mathbf{F}_a \in \mathbb{R}^{k \times D}$ for $\mathcal{N}_a$ using a 3-layer Multi-Layer Perceptron (MLP) where $f_i$ denotes a per-point feature and $D$ is the feature dimension of $f_i$. 
    \textbf{2)} We define a Temperature-adjusted Multi-Head Self-Attention (TMHSA) module and apply it on $\mathbf{F}_a$ extracted in the first step. The TMHSA softly mingles per-point information from different aspects and outputs an integrated feature set $\mathbf{F}' \in \mathbb{R}^{k \times D}$. This is the key step that enables soft neighbourhood selection, points information integration, and attention strength adjustment. 
    \textbf{3)} A Feed Forward Network (FFN) that is introduced in \cite{vaswani2017attention} is applied to transform $\mathbf{F}'$ to $\mathbf{F}'' \in \mathbb{R}^{k \times D}$ in order to introduce non-linearity. 
    \textbf{4)} We apply a point-wise max pooling $g: \mathbb{R}^{k \times D} \rightarrow \mathbb{R}^D $, a symmetric function introduced in PointNet~\cite{PointNet_Charles2017}, on $\mathbf{F}''$ to extract a patch descriptor $\mathbf{d}_a$ for $\mathcal{N}_a$. This patch descriptor is designed to capture the geometric information of this local patch and 
    \textbf{5)} a number of fully-connected (FC) layers are applied to the $\mathbf{d}_a$ to finally regress a normal estimation $\mathbf{\Tilde{n}}_a$ from $\mathbf{d}_a$. 
    With this pipeline, our model learns to estimate normals for different surface properties and to adjust the softness in the self-attention. The full process is illustrated in Fig.~\ref{fig:method_pipeline}.
    
\textbf{Training Loss}
Our network is trained by minimising the unoriented angle difference between estimated normals $\mathbf{\Tilde{n}}$ and ground truth normals $\mathbf{n}$. We use the sine distance $\sin(\mathbf{n}, \mathbf{\Tilde{n}})$ to measure the unoriented angle difference:
    \begin{equation}
        L := \sin(\mathbf{n}, \mathbf{\Tilde{n}}) = \frac{\|\mathbf{\Tilde{n}} \times \mathbf{n}\|}{\|\mathbf{\Tilde{n}}\|\|\mathbf{n}\|}
    \end{equation}

\textbf{TMHSA Module}
The key component of our network architecture is the TMHSA module, which enables our network to decide which neighbours are useful and how they should be combined. We first introduce the Temperature-adjusted Self-Attention (TSA) module, the building block of the TMHSA module. The TSA takes a set of per-point features around a point as input and produces a set of weighted per-point features. Features that are potentially useful to the task are assigned with higher weights, based on the dot product between features~\cite{vaswani2017attention}. Specifically, we define TSA as follows:
    \begin{equation}
        \text{TSA}(\mathbf{F}) := \text{softmax}(\frac{\mathbf{F}_t\mathbf{W}_q \mathbf{W}_k^\top\mathbf{F}_t^\top}{\sqrt{D}})\mathbf{F}_t\mathbf{W}_v
    \end{equation}
where $\mathbf{W}_{q,k,v} \in \mathbf{R}^{D \times D}$ are learnable linear projections that transform $\mathbf{F}$ to other feature spaces and $\mathbf{F}_t = \mathbf{F}/t$ is the temperature-adjusted feature set for a local point patch. In the TSA module, the softmax function produces weights for each feature vector and the temperature $t$ controls the smoothness of the softmax function, for example, the softmax gets sharper when the temperature decreases. By setting $t$ to a learnable parameter, the network can control the sharpness of the softmax function and we denote this behaviour as attention strength adjustment.

The linear transformations $\mathbf{W}_{q,k,v}$ enable the TSA module to attend to one aspect of $\mathbf{F}$. By stacking more linear transformations, we can build the THMSA module that attend to multiple aspects of $\mathbf{F}$:
    \begin{equation}
        \text{TMHSA}(\mathbf{F}) := \text{concat}_{h \in H}[\text{TSA}_h(F)]\mathbf{W}_o
    \end{equation}
where $H$ is the number of heads or the number of linear transformations we apply to $\mathbf{F}$ and $\mathbf{W}_o$ is a learnable projection matrix that combines outputs from all attention heads. 

\textbf{Implementation}
Our network is implemented using PyTorch \cite{paszke2019pytorch}. In the pre-processing stage, we scale all objects to a unit sphere. For each point $a$, we find its $k$ nearest neighbour and shift this $k$-point patch to the origin by subtracting the mean of the patch. We use the Adam optimiser with a start learning rate of 5e-4, a momentum set to 0.9 and no weight decay. The learning rate is scheduled to divide by 10 at epoch 400 and 800 and we train the model for 900 epochs in total. We use a batch size of 12000 and found no significant performance difference with other batch sizes. The convolutional layers are initialised using Kaiming Uniform~\cite{he2015delving} and the temperature parameter is initialised to 1.0.

\section{Experiments}\label{sec:experiments}
In this section, we demonstrate two benefits of our neural network: 1) our method is much less sensitive to the neighbourhood range in comparison with other approaches. 2) our method achieves state of the art performance with a much smaller model and a much faster inference time. We organise this section as follows: Sec.~\ref{subsec:dataset_and_metrics} introduces the dataset details and evaluation metrics to compare with other approaches. Sec.~\ref{subsec:full_results} presents our main results on stability, performance, model size, and inference time. Sec.~\ref{subsec:discussion} discusses the reason why our model is insensitive to the neighbourhood range variance using attention maps. To validate that learning a temperature parameter is helpful, we provide an ablation test in Sec.~\ref{subsec:ablation}. Lastly, Sec.~\ref{subsec:icp_application_study} provides an application study that uses the ICP algorithm as an example to show how downstream tasks can benefit from the neighbourhood-insensitive property.

\subsection{Dataset and Metrics} \label{subsec:dataset_and_metrics}
We train our model using the PCPNet dataset \cite{guerrero2018pcpnet} that consists of 27 objects with each object containing 100,000 points. The training/validation/test splits contain 8/3/19 objects respectively. We do not use the noise-augmented data provided in the PCPNet dataset and we follow the train/test splits of PCPNet \cite{guerrero2018pcpnet} and Nesti-Net \cite{ben2019nesti} to ensure a fair comparison.

We evaluate our normal estimation network using root mean square error (RMSE) and the proportion of good points (PGP$\alpha$) metric, the same as Nesti-Nets\cite{ben2019nesti} and PCPNet\cite{guerrero2018pcpnet}. Both RMSE and PGP are computed using the unoriented angle error $\beta$. In the definition of unoriented angle error, a $179\degree$ angle difference is considered as accurate as a $1\degree$ angle difference. We compute the unoriented angle error using 
        $\beta = \arccos\left(\big| \frac{\mathbf{\Tilde{n}} \cdot \mathbf{n}}{\|\mathbf{\Tilde{n}}\|\|\mathbf{n}\|} \big|\right)$
where $\mathbf{\Tilde{n}}$ is a predicted normal vector and $\mathbf{n}$ is a ground truth vector. PGP$\alpha$ computes the percentage of angle differences between the predicted result and ground truth is less than $\alpha\degree$, we compute the PGP$\alpha$ using 
        $\text{PGP}\alpha = \frac{1}{N} \sum_{a \in P} \mathbbm{1}(\beta_a<\alpha)$
, where $\mathbbm{1}(\cdot)$ is an indicator function that yields 1 when the condition inside the operator holds and $N$ is the cardinality of the point cloud $P$.

    \begin{figure}[]
        \centering
        \includegraphics[width=0.9\linewidth]{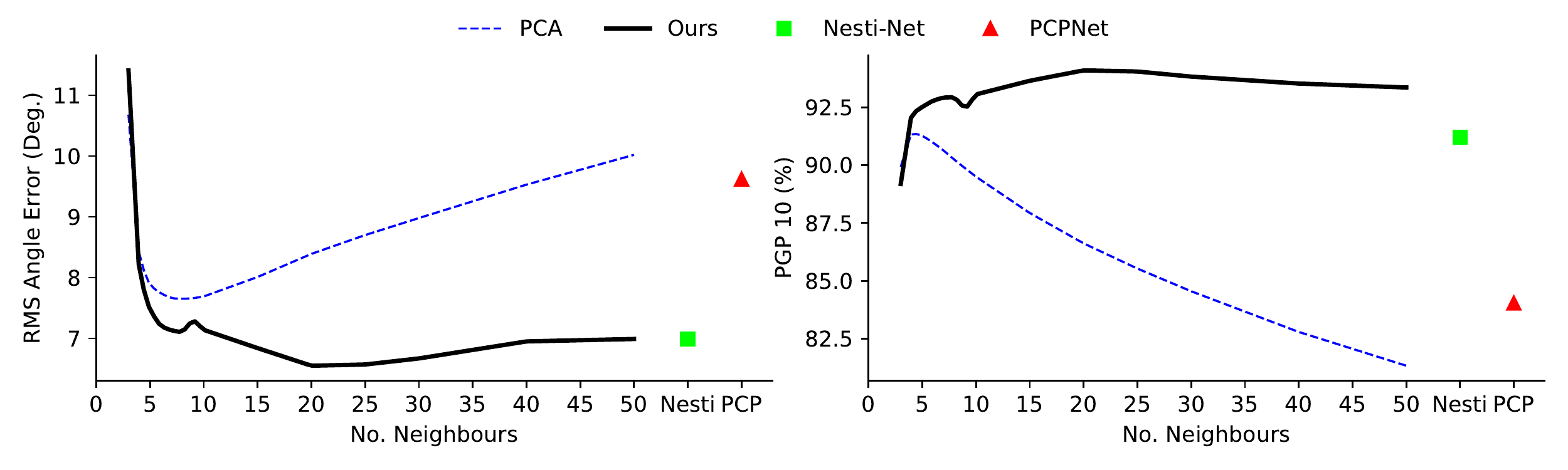}
        \caption{Model performance comparison, given the different number of neighbours. Nesti-Net and PCPNet are plotted on the side since they take multi-scale patches as input.} 
        \label{fig:ours_vs_others}
    \end{figure} 

\subsection{Results} \label{subsec:full_results}
\textbf{Stability} We show that our network is insensitive to the range of neighbourhood in Fig.~\ref{fig:ours_vs_others}. Our model outperforms PCA by a large margin and most importantly, the performance of our method does not degrade as the number of neighbours increases. Nesti-Net~\cite{ben2019nesti} and PCPNet~\cite{guerrero2018pcpnet} are shown separately in this figure because they use patches from multi-scales. 

Our method consistently performs better than PCA and PCPNet as long as more than 3 neighbour points are used, and better than Nesti-Net when more than 15 neighbours are available. PCA achieves the best performance using 7 or 8 neighbours and our model outperforms the top PCA performance when more than 5 neighbours are given. This performance is desirable for downstream algorithms because it ensures that high-quality normal estimations could be found by simply setting a large enough neighbourhood range, and our network can choose neighbours it needs adaptively based on local geometric structures.

\textbf{State of the Art Performance} In addition to achieving a neighbourhood-insensitive property, equipped with the attentional neighbourhood integration model, our method also outperforms other methods by a large margin with a much smaller model and a faster inference time (Tab.\ref{tab:sota_perf}).

    \begin{table}[h]
    \centering
    \begin{tabular}{lccc}
    \hline
    Models    & PGP 5 (\%)     & PGP 10 (\%)    & RMSE (deg)    \\ \hline
    PCA~\cite{hoppe1992surface}       & 81.84          & 90.23          & 7.65          \\
    Jet~\cite{cazals2005estimating_jet}       & 79.05          & 88.02          & 7.60          \\
    PCPNet~\cite{guerrero2018pcpnet}    & 69.86          & 84.04          & 9.62          \\
    Nesti-Net~\cite{ben2019nesti} & 80.57          & 91.20          & 6.99          \\ \hline
    Ours      & \textbf{86.24} & \textbf{94.10} & \textbf{6.55} \\ \hline
    \end{tabular}
    \caption{Normal estimation results on PCPNet dataset. Our model achieves the state of the art results in all metrics.}
    \label{tab:sota_perf}
    \end{table}

\textbf{Model Size and Inference Time} As mentioned above, our model is much smaller and faster than the previous state of the art model Nesti-Net~\cite{ben2019nesti}, which has 8 sub-networks (7 experts and a manager). Our model takes 41.1MB and Nesti-Net takes 1051.6MB (25x larger). Regarding inference time, our model takes 6.5 seconds to evaluate 100,000 normals (50 neighbours for each normal), while the multi-scale Nesti-Net takes 81.1 seconds. We also report that PCA consumes 0.27 seconds given 8 neighbours, where PCA produces best normal estimations according to Fig.~\ref{fig:ours_vs_others}. For learning-based models, we benchmark timings on an NVIDIA-1080Ti GPU. For PCA, we use NumPy's SVD function and an Intel i9-9900K CPU. Detailed model sizes and inference time are listed in Tab.~\ref{tab:model_size_time}.
    \begin{table}[h]
    \centering
    \begin{tabular}{lcc}
    \hline
    Model     & Model Size(MB) & \begin{tabular}[c]{@{}c@{}}Inference Time \\ (sec / 100K points)\end{tabular} \\ \hline
    PCA~\cite{hoppe1992surface}       & N/A            & 0.27 (nb8)                                                                    \\
    PCPNet~\cite{guerrero2018pcpnet}    & 90             & 165.5                                                                         \\
    Nesti-Net~\cite{ben2019nesti} & 1052           & 81.1                                                                          \\
    Ours      & 41             & 6.5 (nb50)                                                                    \\ \hline
    \end{tabular}
    \caption{Model sizes and mean inference time. Our model is 25x times smaller and 12x faster than Nesti-Net. The time shown in the table only accounts for forward passing time, i.e. no data loading, neighbour searching, pre-processing and post-processing time included. PCPNet is slow because it learns a quaternion transformation and converts it to a rotation matrix in an inefficient way. nb denotes the number of neighbours used in normal estimation.}
    \label{tab:model_size_time}
    \end{table}

\subsection{Ablation Study} \label{subsec:ablation}
\begin{figure}[h]
    \centering
    \includegraphics[width=0.9\linewidth]{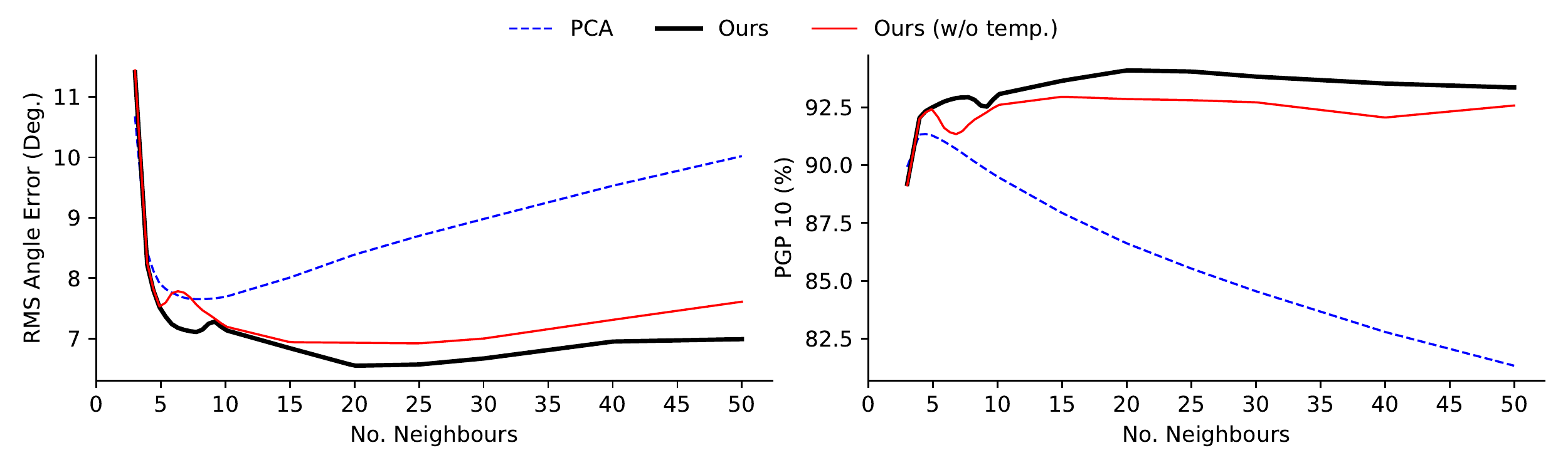}
    \caption{Ablation study on the learnable temperature. Learning a temperature parameter is beneficial to improving stability and performance.} \label{fig:w_vs_wo_temp}
    \end{figure}

We verify that introducing a temperature parameter improves the network performance and makes the network more robust to large neighbourhood ranges in Tab.~\ref{tab:abl_temperature} and Fig.~\ref{fig:w_vs_wo_temp}. By learning a temperature, our network learns to control the softmax in the TMHSA module and to control the attention strength.
    
    \begin{table}[h]
    \centering
    \begin{tabular}{lccc}
    \hline
    Model           & PGP 5 (\%) & PGP 10 (\%) & RMSE (deg) \\ \hline
    w/o temperature & 85.50      & 92.81       & 6.92       \\
    w temperature   & \textbf{86.24}      & \textbf{94.10}       & \textbf{6.55}       \\ \hline
    \end{tabular}
    \caption{Ablation study on learning the temperature parameter.}
    \label{tab:abl_temperature}
    \end{table}

\subsection{Attention Weights Analysis} \label{subsec:discussion}
\begin{figure}[t]
    \centering
    \includegraphics[width=0.7\linewidth]{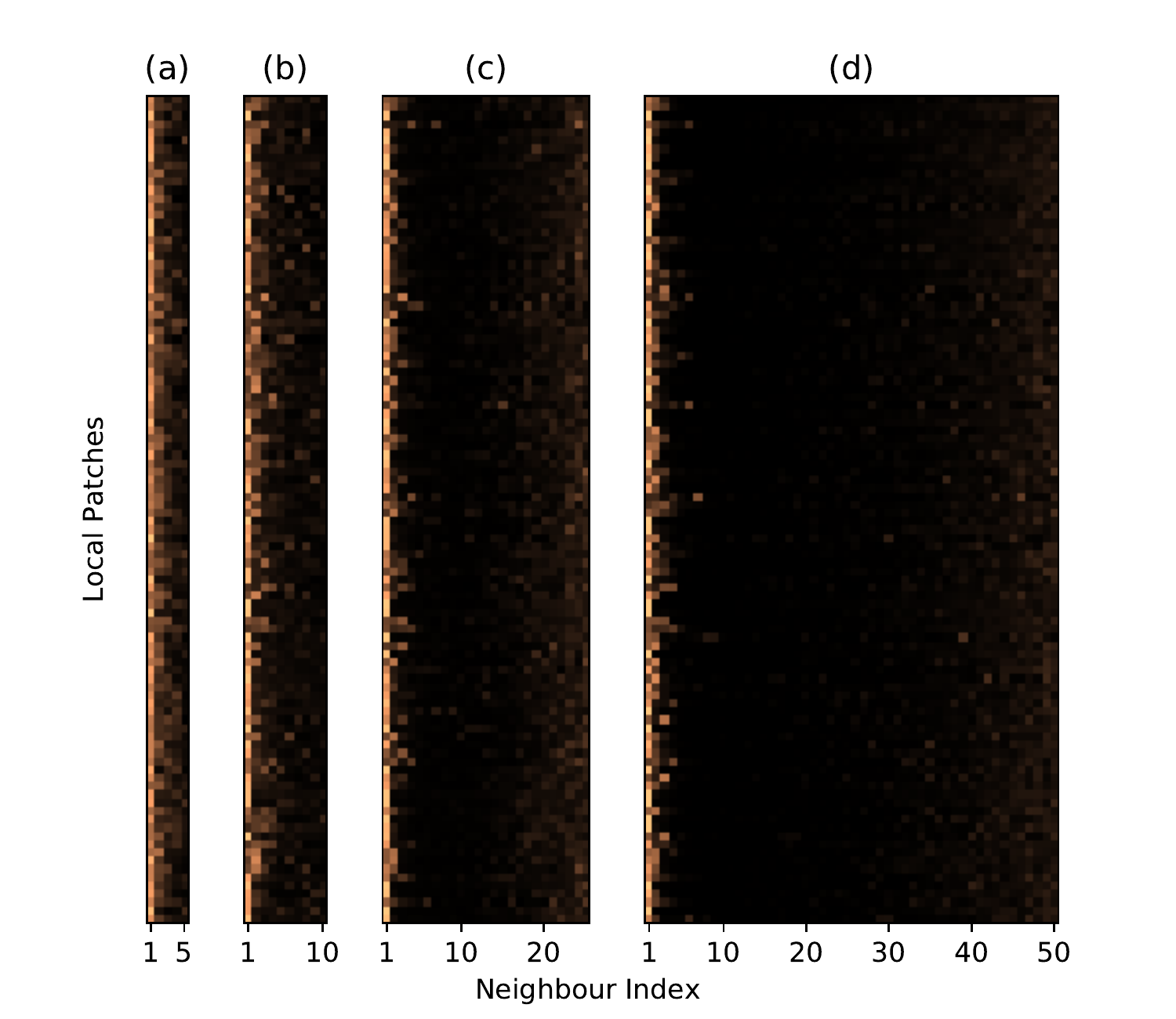}
    \caption{Attention weights visualisation. Each row in the attention map is a local patch that contains 5, 10, 25, 50 neighbours, from to left to right. The x-axis denotes the indices of closest neighbours within a patch, i.e., the point that is the closest to the patch centre is represented by the $1^{st}$ pixel in a row.}
    \label{fig:attn-map}
\end{figure}

In this section, we discuss why the performance of our network does not degrade when the number of neighbours are increasing, as in PCA. In fact, our network performs better when more neighbours are available. To explain this, we show 4 attention maps from our models in Fig.~\ref{fig:attn-map}. The four models we use to generate these attention maps are trained and tested using local patches contain 5, 10, 25, and 50 neighbours and achieves PGP10 92.5\%, 93.0\%, 94.0\%, and 93.3\%, from left to right, while the best PCA performance is 91.4\%. In an attention map, each local patch is represented by a row of pixels and attention weights are colour-coded by pixel intensity.

\begin{figure}[h!]
    \centering
    \includegraphics[width=0.7\linewidth]{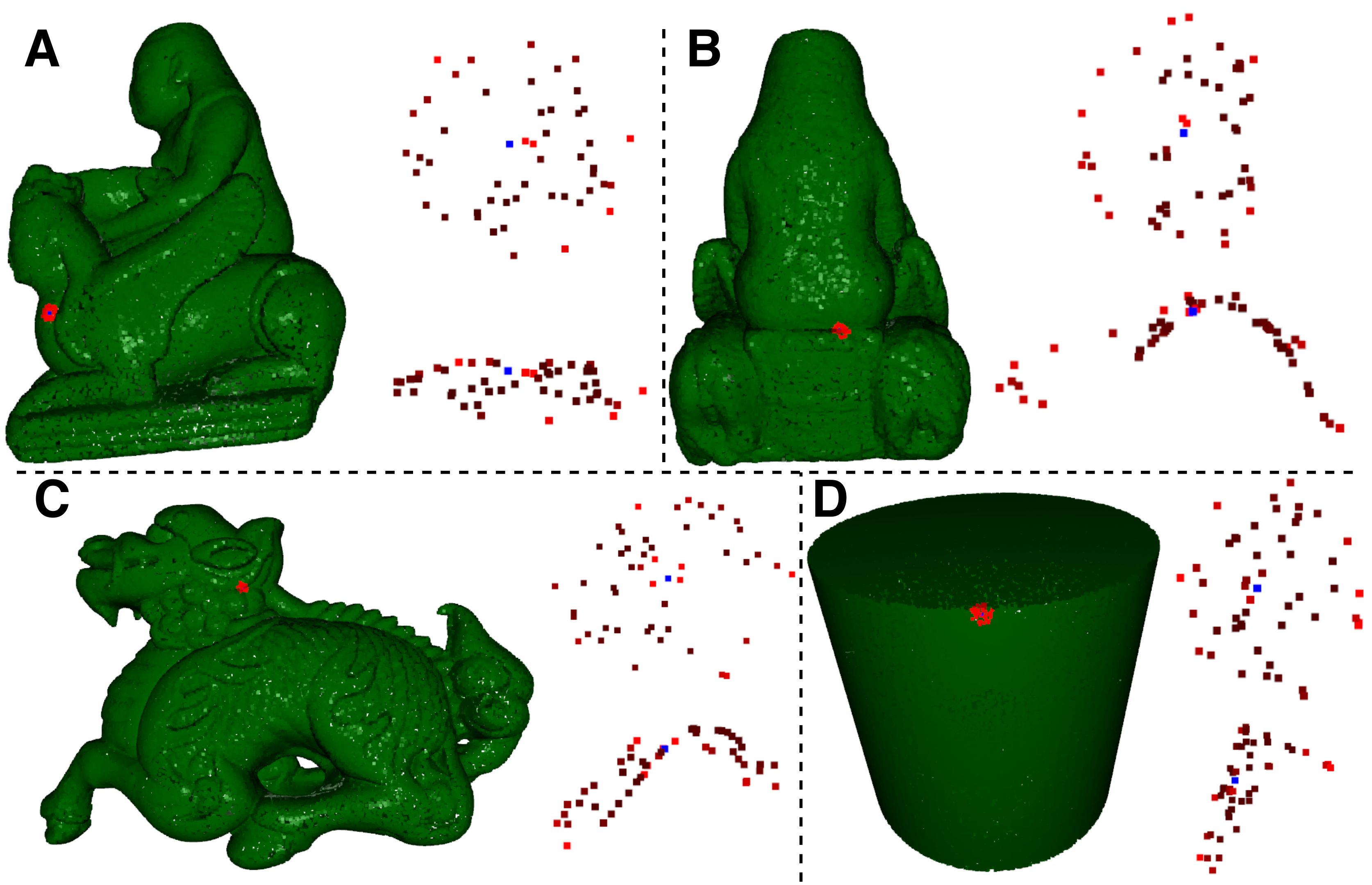}
    \caption{Attention weights on various patches evaluated on the test set. The point of our interest $p$ is coloured with \textcolor{blue}{blue} and its 49 neighbours are coloured with \textcolor{red}{red}. For each patch, we first highlight its points within the entire point cloud, then we show two colour-coded views of the patch, in which brighter red denotes higher attention weight. Patch A is at a flat area while patch B, C, and D contain more complicated geometry structures. The figure is best viewed when zoomed in. We also attached a high resolution version in the supplementary material.}
    \label{fig:attn_vis_3d}
\end{figure}

From the attention map over 5 and 10 neighbours, (a) and (b), we can see that the network pays more attention to neighbours that are close to the patch centre and ignores points farther than $8^{th}$ neighbours. This agrees with the PCA performance in Fig.~\ref{fig:ours_vs_others}. From the attention map over 25 and 50 neighbours, (c) and (d), we observe that the network pays extra attention to points at the right-end in a row. These are points far from the patch centre. To illustrate this better, we show the attention weights predicted by our network in 3D in Fig. \ref{fig:attn_vis_3d}. These attention weights visualisations suggest that the network learns to identify geometric properties around the current point using points from larger scales, by inspecting, for example, whether the patch contains a sharp edge or a corner.  Therefore, our network maintains a relatively stable performance despite different numbers of neighbours by focusing on relevant points.

\subsection{Application Study: ICP Point Cloud Registration} \label{subsec:icp_application_study}
In this section, we demonstrate how other downstream algorithms can benefit from our neighbourhood-insensitive normal estimation network. We take the point-to-plane ICP ~\cite{ICP_Besl1992, chen1992object} algorithm as an example and show that, by simply setting the number of neighbours $k$ to a large enough number, for example, $k=50$ in our network, ICP with our normal estimation converges faster than with other normal estimation methods. Moreover, ICP with classical normal estimation methods like PCA requires fine-tuning $k$ manually to achieve best performance, which is still outperformed by our network.

\textbf{ICP Preliminary} 
ICP is a widely used point cloud registration and pose estimation method. Many variants have been proposed in past decades. The original point-to-point ICP~\cite{ICP_Besl1992} optimises the distance between closest point pairs directly. The point-to-plane ICP~\cite{chen1992object} is proposed to improve the converging speed and robustness against noise. Given two point clouds, a source point cloud $\mathbf{X}_s$ and a rigid transformed destination point cloud $\mathbf{X}_d = \mathbf{T}_{ds} \mathbf{X}_s$, ICP computes the rigid transformation $\mathbf{T}_{ds} \in \mathbb{SE}(3)$ to register $\mathbf{X}_d$ and $\mathbf{X}_s$ iteratively. The point-to-point energy is defined as
    \begin{equation}
        E_{pt-pt} = \sum(\|\mathbf{x}_s - \mathbf{x}_d\|^2)
        \label{eq:icp-ptpt}
    \end{equation}
and the point-to-plane energy is defined as
    \begin{equation}
        E_{pt-plane} = \sum(\mathbf{n}_d \cdot \mathbf{(\Tilde{T}}_{ds} \mathbf{x}_s - \mathbf{x}_d))^2
        \label{eq:icp-ptplane}
    \end{equation}
where $\mathbf{\Tilde{T}}_{ds}$ denotes the estimated transformation, $\mathbf{x}_s$ is a point in $\mathbf{X}_s$, $\mathbf{x}_d$ is the corresponding point in $\mathbf{X}_d$, and $\mathbf{n}_d$ is the surface normal at $\mathbf{x}_d$. 

\textbf{Results}
We run the point-to-plane ICP on the test-set of PCPNet dataset~\cite{guerrero2018pcpnet}, using normals from different estimation methods (Tab.~\ref{tab:icp_iters}). With normal estimation from our model, the point-to-plane ICP converges faster than using normals from PCA~\cite{hoppe1992surface} and Nesti-Net~\cite{ben2019nesti}. By making comparison between the nb25 and nb50 models at the last two rows in Tab.~\ref{tab:icp_iters}, we can conclude our model is stable given the different number of neighbours. For PCA, we run an exhaustive search over the different numbers of neighbours and report the smallest iteration number in Tab.~\ref{tab:icp_iters} (the PCA-full row). The full ICP iteration table for PCA and the full shape names s1-s19 can be found in the supplementary material.

    \renewcommand{\tabcolsep}{2pt}
    \begin{table}[]
    \centering
    \resizebox{\textwidth}{!}{
        \begin{tabular}{cccccccccccccccc}
        Model & s01 & s02 & s03 & s04 & s07 & s08 & s09 & s10 & s11 & s13 & s14 & s15 & s16 & s19 & Mean \\ \hline
        GT & 7 & 16 & 6 & 10 & 6 & 7 & 9 & 11 & 7 & 26 & 10 & 10 & 6 & 22 & 10.9 \\ \hline
        PCA-nb8 & 15 & 52 & 13 & 23 & 20 & 13 & 10 & 24 & \textbf{9} & 37 & 64 & 16 & \textbf{7} & F & 23.3 \\
        PCA-full & 14(4) & 31(4) & 12(4) & 16(4) & 14(15) & 10(3) & \textbf{9(7)} & 14(7) & \textbf{9(3)} & 33(3) & 17(20) & 13(3) & \textbf{7(3)} & \textbf{12(15)} & 15.1(6) \\
        Nesti-Net & 35 & F & 11 & F & 16 & 13 & 12 & 84 & F & F & F & F & F & F & 28.5 \\ \hline
        Ours-nb50 & \textbf{8} & \textbf{17} & \textbf{8} & \textbf{10} & \textbf{7} & \textbf{8} & \textbf{9} & \textbf{11} & 9 & \textbf{29} & \textbf{12} & \textbf{12} & 7 & F & \textbf{11.3} \\
        Ours-nb25 & \textbf{8} & \textbf{17} & \textbf{8} & \textbf{10} & \textbf{7} & \textbf{8} & 10 & 16 & \textbf{8} & \textbf{29} & 13 & \textbf{12} & \textbf{6} & F & 11.7 \\ \hline
        \end{tabular}
    }
    \caption{The number of iterations that the point-to-plane ICP takes to converge, using normals from different sources. s01-s19 denotes the 19 shapes in the test-set of PCPNet~\cite{guerrero2018pcpnet} dataset. Shapes s05-06, s12, and s17-18 are not shown because ICP fails to converge even using ground truth normals. We denote failure to converge with F. The number in parentheses is the number of neighbours that produces the smallest number of iterations for PCA, for the specific shape. GT denotes ground truth normals. PCA-nb8 denotes the PCA result using the best global $k=8$ (according to Fig.~\ref{fig:ours_vs_others}). nb25 and nb50 denote models trained with 25 and 50 neighbours.}
    \label{tab:icp_iters}
    \end{table}

\textbf{ICP Setup}
To ensure fair comparisons, we set an early stopping criteria using the point-to-point energy $E_{pt-pt}$ since this stopping criterion is immune to the normal quality when evaluating registration results. A perfect registration should produce $E_{pt-pt} = 0$ but the $E_{pt-plane}$ can still be larger than zero because the normals estimations are not perfect. We set the stopping threshold $E_{pt-pt}^{stop}$ to $10^{-5}$. For simplicity, we set the rotation angles between $\mathbf{X}_s$ and $\mathbf{X}_d$ to $10\degree$ in all three axes ($x, y, z$), and the translation between $\mathbf{X}_s$ and $\mathbf{X}_d$ to $[0.01, 0.01, 0.01]$. We scale all point clouds to a unit sphere and shift them to the point cloud centre before running ICP. We use the Levenberg-Marquardt~\cite{Levenberg1944, Marquardt1963} solver to optimise $E_{pt-plane}$. Our ICP code comes along with the training code at \url{https://code.active.vision}.

\section{Conclusion}\label{sec:conclusion}
In this work, we propose an attention-based normal estimation network that is insensitive to neighbourhood range and provides state of the art normal estimation performance. We achieve this by applying an attention module to the local neighbourhood so our network can focus on relevant points and integrate information softly. With a learnable temperature parameter, our network can also control how attentive it should be under current neighbourhood settings. As a result, our network is 12x faster and 25x smaller than the previous state of the art models~\cite{ben2019nesti} and provides 2.9\% higher accuracy (PGP 10) on the PCPNet dataset~\cite{guerrero2018pcpnet}. We also use point-to-plane (ICP) as an application to show that our normal estimations lead to faster convergence than normal estimations from other methods, without manually fine-tuning neighbourhood range parameters.

\section*{Acknowledgement}
The authors would like to thank Min Chen, Tengda Han, Shuda Li, Tim Yuqing Tang and Shangzhe Wu for insightful discussions and proofreading.

\bibliography{references.bib}

\end{document}